\documentclass[letterpaper]{article} 
\usepackage{aaai24}  
\usepackage{times}  
\usepackage{helvet}  
\usepackage{courier}  
\usepackage[hyphens]{url}  
\usepackage{graphicx} 
\usepackage{natbib}  
\usepackage{caption} 
\usepackage{algorithm}
\usepackage{algorithmic}
\usepackage{subfigure}
\usepackage{amsfonts}
\usepackage{amsmath}
\usepackage{amssymb}
\usepackage{booktabs}
\usepackage{tikz}
\usepackage{multirow}
\usepackage{dsfont}
\usepackage{enumitem}
\usepackage{colortbl}
\definecolor{table_yellow}{RGB}{254,243,222}
\definecolor{table_blue}{RGB}{237,245,255}
\definecolor{table_gray}{RGB}{245,245,245}
\usepackage{newfloat}
\usepackage{listings}
\DeclareCaptionStyle{ruled}{labelfont=normalfont,labelsep=colon,strut=off} 
\floatstyle{ruled}
\newfloat{listing}{tb}{lst}{}
\floatname{listing}{Listing}
\title{MELO: Enhancing Model Editing with Neuron-Indexed Dynamic LoRA}
\author{
    Lang Yu\textsuperscript{\rm 1,2},
    Qin Chen \textsuperscript{\rm 1,2}\thanks{Corresponding author: Qin Chen (qchen@cs.ecnu.edu.cn)},
    Jie Zhou \textsuperscript{\rm 1,2},
    Liang He \textsuperscript{\rm 1,2}
}
\affiliations{
    \textsuperscript{\rm 1}School of Computer Science and Technology, East China Normal University\\
    \textsuperscript{\rm 2}Shanghai Institute of AI for Education, East China Normal University\\
    lyu@stu.ecnu.edu.cn, \{qchen, jzhou, lhe\}@cs.ecnu.edu.cn
}

\usepackage{bibentry}

\begin{document}

\maketitle

\begin{abstract}
Large language models (LLMs) have shown great success in various Natural Language Processing (NLP) tasks, whist they still need updates after deployment to fix errors or keep pace with the changing knowledge in the world. Researchers formulate such problem as Model Editing and have developed various editors focusing on different axes of editing properties. However, current editors can hardly support all properties and rely on heavy computational resources. In this paper, we propose a plug-in Model Editing method based on neuron-indexed dynamic LoRA (MELO), which alters the behavior of language models by dynamically activating certain LoRA blocks according to the index built in an inner vector database. Our method satisfies various editing properties with high efficiency and can be easily integrated into multiple LLM backbones. Experimental results show that our proposed MELO achieves state-of-the-art editing performance on three sequential editing tasks (document classification, question answering and hallucination correction), while requires the least trainable parameters and computational cost. 

\end{abstract}

\section{Introduction}

\begin{figure}[hbt!] \centering
    \includegraphics[width=0.45\textwidth]{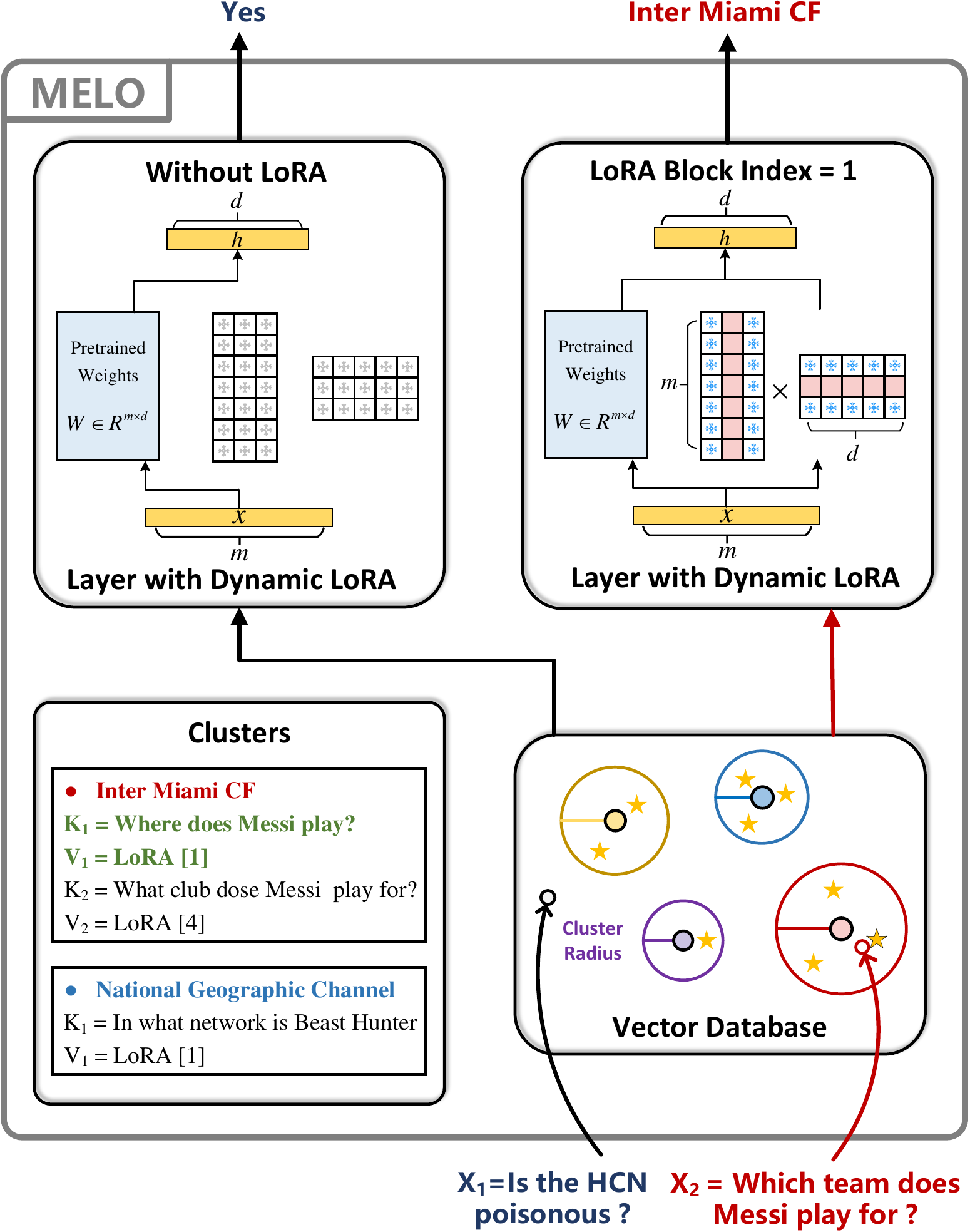}
    \caption{MELO integrates dynamic LoRA modules into LLMs, which are indexed in an inner vector database. During training, the edits are learned with non-overlapping LoRA blocks. In the inference phase, the inputs $X_1$ and $X_2$ are searched in the vector database, and certain LoRA blocks (or none) are activated for post-edit response.} \label{intro}
\end{figure}

With well-designed architectures and ever-growing size, large language models (LLMs) \cite{gpt3,llama} have become the paradigm for solving many Natural Language Processing (NLP) tasks. However, they still need updates after deployment to calibrate hallucination and keep pace with the changing knowledge over time. Meanwhile, it’s infeasible to frequently re-train or fine-tune LLMs on upstream datasets due to high computational cost. This indicates a need to develop editors enabling effective but cheap updates for large pre-trained models.

Researchers formulate such problem as Model Editing \cite{edit_survey} and have proposed various editors focusing on different axes of editing properties. Prior studies MEND and SERAC \cite{mend,serac} primarily define the fundamental properties \textbf{Edit Success} and \textbf{Locality}, which require effective updates to LLMs within a domain of interest, while ensure no performance degradation on other inputs. Whereas, their work relies on  extra training data for editing. ROME and MEMIT \cite{rome,memit} support large-scale direct edits by locating knowledge in specific layers of GPT, and further achieves \textbf{Generality} for associated inputs, yet the inputs are restricted to the directional $(s,r,o)$ relations. Recent studies GRACE \cite{grace} and T-Patcher \cite{patcher} investigate \textbf{Sequential Editing} for streaming edits, which utilize external memory of hidden states or neurons to solve catastrophic forgetting, but large amount of training time and computational resources are required for extensive edits. Despite the promising progress, previous methods can hardly achieve all editing properties with high resource efficiency.

In this paper, we propose MELO\footnote{Code is available at https://github.com/BruthYU/MELO}, which performs \textbf{M}odel \textbf{E}diting with neuron-indexed dynamic \textbf{Lo}w-rank adapter. As shown in Figure \ref{intro}, MELO alters the behavior of language models by dynamically activating certain blocks of low-rank adapter (LoRA) according to the index built in an inner vector database. Furthermore, it could support all editing properties as follows:  
 
\begin{itemize}
    \item[(1)] \textbf{Edit Success}: Each batch of edits is trained with a unique set of LoRA blocks, which will be accurately invoked during inference for in-scope inputs.
    \item[(2)] \textbf{Locality}: An inner vector database is built to identify the editing scope, hence the inputs out of the scope will retain original predictions. 
    \item[(3)] \textbf{Generality}: Semantic clusters with different radii are built for covering the associated edits. 
    Corresponding LoRA blocks will be activated once the input falls in the scope of one cluster.
    \item[(4)] \textbf{Sequential Editing}: Sequential batches are trained with non-overlapping LoRA blocks, which addresses the issue of catastrophic forgetting on previous edits.
    \item[(5)] \textbf{Efficiency}: MELO merely employs dynamic LoRA blocks with small partial rank for editing, which can learn large batches of edits with very few parameters. 

\end{itemize}

We perform experiments on three well-known editing tasks, namely document classification, question answering and hallucination correction, and the results demonstrate the great advantages of our proposed method. The main contributions of our work can be summarized as follows: 
\begin{itemize}
\item  We propose a plug-in model editing method with neuron-indexed dynamic LoRA, which alters models' behavior by activating corresponding LoRA blocks, and can be seamlessly integrated into various LLM backbones. 
\item We explore the potential of vector database to memorize edits, which well builds the editing scope in the training stage and provides neuron index to find the exact LoRA blocks for post-edit inputs during inference.
\item Extensive experiments on three sequential editing tasks indicate that our proposed method achieves the state-of-the-art editing performance compared with the recent baselines. In particular, our method well supports all editing properties without using extra training data.

\end{itemize}

\section{Related Work}
Model editing has attracted great attention in recent years \cite{edit_survey}. Existing methods mainly focus on four editing properties (edit success, locality, generality and sequential editing), and can be categorized into three groups: meta-learning editors, locate-then-edit editors and memory-based editors. \textit{Meta-learning editors} employ external network to predict necessary gradient for editing. MEND \cite{mend} learns a hyper-network to transform the gradient obtained by standard fine-tuning, which enables efficient updates to LLMs but needs additional data for training.
As to the \textit{locate-then-edit editors}, they initially identify parameters corresponding to the intended edits and then modify target parameters with direct updates. ROME and MEMIT \cite{rome,memit} propose to locate knowledge in GPT-based models and then modify a sequence of layers to facilitate extensive edits, whereas they are restricted to directional $(s,r,o)$ relations. For \textit{memory-based editors}, the specific hidden states or neurons that store the edit knowledge are used for post-edit response. SERAC \cite{serac} employs a scope classifier and routes inputs to the frozen model or the counterfactual model. CaliNet \cite{calinet} and T-Patcher \cite{patcher} attach neurons for each edit, while GRACE \cite{grace} replaces hidden states of in-scope inputs with parameters searched from a codebook for edit memorization. Whereas, all these methods can hardly achieve all editing properties with high efficiency, which is difficult to adapt to real editing scenarios, especially for models with large-scale parameters. Thus, we aim to explore a more effective and efficient model editing method that satisfies all editing properties.

\subsection{Parameter-Efficient Tuning}
The key idea of parameter-efficient tuning is to insert a tiny trainable module to a large pre-trained model and optimize task-specific losses by only adjusting module parameters. The most representative methods are Adapter, Prompt Tuning and LoRA. Adapter \cite{adapter,bitfit} is a trainable bottle-neck shaped neural network prepended to a transformer block's output. Prompt Tuning \cite{prefix,vpt} aims to adapt pre-trained models to downstream tasks by optimizing appended prompts in the form of discrete tokens or continuous vectors. LoRA \cite{lora,adalora,dylora} keeps the model frozen, and only updates rank decomposition matrices truncated to the target modules. Inspired by DyLoRA \cite{dylora} that randomly updates partial parameters of the LoRA module each time, we propose to index isolated LoRA blocks to efficiently alter models' behavior.  

\subsection{Domain Specialization}
Domain specialization~\cite{domain} is a critical yet challenging problem to enhance the domain-specific expertise of LLMs. Approaches that specialize models with domain knowledge can be categorized into three classes: \textit{(1) External Augmentation} uses external resources or tools~\cite{webgpt,toolformer} to incorporate the domain-knowledge into the input prompt or generated output. \textit{(2) Prompt Crafting} involves discrete~\cite{COT} or learnable prompts~\cite{spot} to activate domain knowledge in pre-trained models. \textit{(3) Model Fine-tuning} updates the LLM's parameters~\cite{lora, dylora} to directly incorporate domain-specific knowledge into the model. In contrast, our proposed MELO could also be used for domain specialization, which could handle scaling number of edits with high efficiency.

\begin{figure*}[t] \centering
    \includegraphics[width=\textwidth,height=0.4\textwidth]{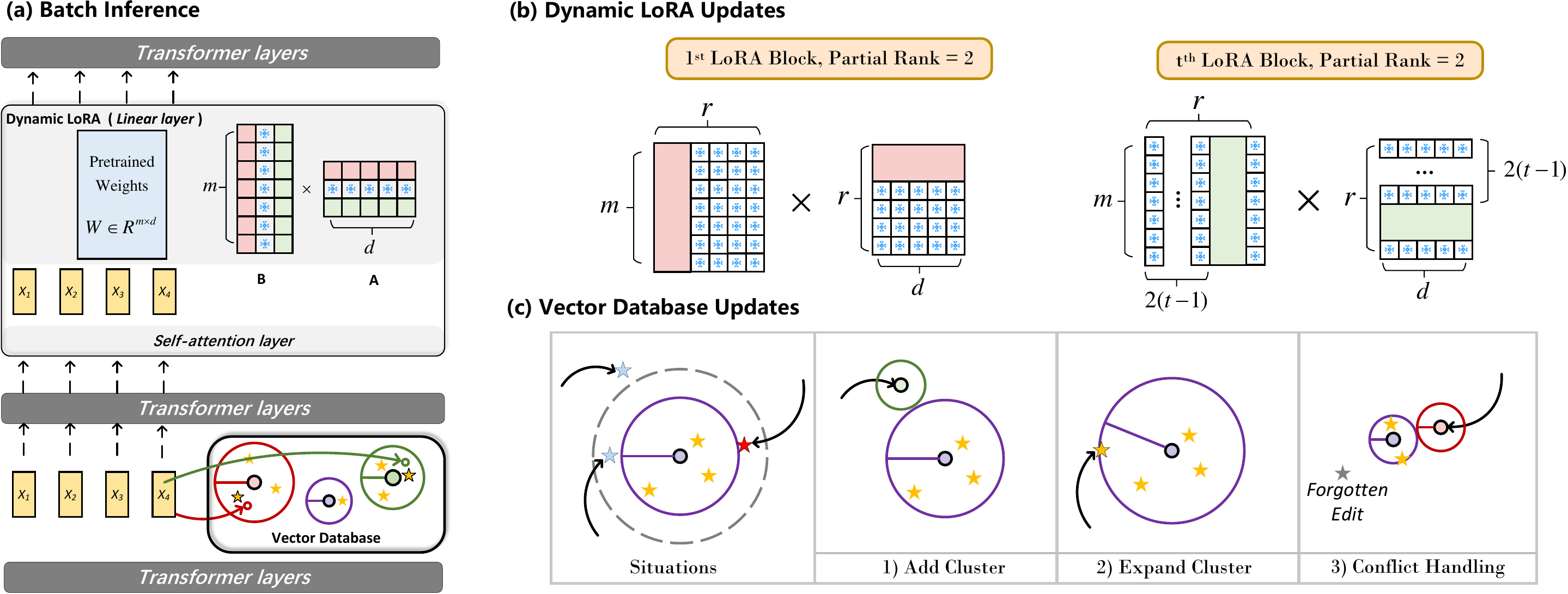}
    \caption{The overall framework of MELO. Each batch of edits is learned in a set of LoRA blocks located in different layers but with the same index. The partial rank of LoRA blocks could be set as a hyper-parameter. Meanwhile, the vector database updates its clusters during training for future LoRA block searching in the inference stage.} \label{main}
\end{figure*}

\section{Method}
 Figure \ref{main} draws the framework of our proposed MELO. The general workflow of the post-edit model is demonstrated in Figure \ref{main}(a): Given a batch of inputs, MELO first searches over the neuron-index built in vector database and then dynamically activate LoRA blocks summed to the original weights, which are trained on associated edits. During the training phase shown in Figure \ref{main}(b) and \ref{main}(c), different batches of edits are trained with non-overlapping LoRA blocks, and the edit samples (key-value pairs) are clustered based on their semantic keys in the vector database, with values indicating the index of LoRA blocks. Details about the editing task and our method are presented as follows.

\subsection{Problem Formulation}
Following the prior works \cite{serac} and \cite{patcher}, we consider the task of editing a base model $f_{base}$ using an dataset $D_{edit} = \{d_1, d_2, ..., d_n\}$ with $n$ sequential batches. Each batch $d_i$ contains several edit input-output pairs $[x_e, y_e]$. 
$R(\cdot)$ denotes a function that rephrases $x_e$ to associated inputs.  
Meanwhile, $[x, y]\in D_{out}$ indicates the samples out of the editing scope. After editing with $t \in [1,n]$ batches of edits, a post-edit model $f_t$ is obtained. During the editing process, a good model editor should meet requirements of the following properties:
\newtheorem{theorem}{Property}
\begin{theorem}
    \textbf{Edit Success}: The model $f_t$ should output desired predictions on intended edits:
    \begin{equation}
        f_t(x_e) = y_e, \forall x_e\in d_{1:t}
    \end{equation}
\end{theorem}

\begin{theorem}
    \textbf{Locality}: The model $f_t$ should retain original predictions on inputs out of the editing scope:
    \begin{equation}
        f_t(x) = f_{base}(x), \forall x\in D_{out}
    \end{equation}
\end{theorem}

\begin{theorem}
    \textbf{Generality}: The model $f_t$ should be able to generalize edits over other equivalent inputs:
    \begin{equation}
        f_t[R(x_e)] = f_t(x_e) , \forall x_e\in d_{1:t}
    \end{equation}
\end{theorem}

\begin{theorem}
    \textbf{Sequential Editing}: The model $f_t$ should align with $f_{t-1}$ on the different set $d_{1:t-1}-d_t$. For recurring edits with new labels $y_e^{t}$, the latest one shall prevail:
    \begin{equation}
        f_t(x_e) = \{
        \begin{aligned}
        &\ f_{t-1}(x_e)&,\forall x_e\in d_{1:t-1}-d_t \\
        &\ y_e^t&,\forall x_e\in d_{1:t-1}\cap d_t
        \end{aligned}
    \end{equation}
\end{theorem} 
Additionally, the \textbf{Property 5 \emph{Efficiency}}  is another requirement for model editors to make pre-trained LLMs quickly adaptable on edits with light computational cost.
\subsection{LoRA: Low-rank Adapter}
We first make a review of the vanilla LoRA techniques \cite{lora}, which hypothesize the updates to any weights have a low ``intrinsic rank". With LoRA, some chosen layers in a frozen LLM are summed with parallel low-rank adapter modules. During fine-tuning, only the LoRA modules can be updated. Assume that $W_0 \in \mathds{R}^{m\times d}$ is a pre-trained weight matrix in model which is accompanied by a LoRA decomposition $\Delta{W} = BA$, where $B\in \mathds{R}^{m\times r}$, $A\in \mathds{R}^{r\times d}$ and $r\ll min(m,d)$. For original $h = W_0 x$, the modified forward pass yields:
\begin{equation}
\label{lora_eq}
    h = W_0 x + \Delta{W} x = W_0 x + \frac{\alpha}{r} BAx
\end{equation}
where $\alpha$ is a constant hyper-parameter for scaling, $B$ is initialized as a zero matrix and $A$ is initialized using a zero-mean Gaussian distribution.

To demonstrate the usage of vanilla LoRA in model editing, we can simply assume that there is only one LoRA module in the pre-trained network. Let's consider a general loss function $\mathcal{L}$ of model $f$ to be edited, the target matrices $B^\star$ and $A^\star$ trained on batch $ d_t = (X_e^t,Y_e^t)$ are formulated as: 
\begin{equation}
    B^\star,A^\star = \mathop{\arg\min}\limits_{B,A} \mathcal{L}[f(X_e^t;BA),Y_e^t]
\end{equation}
where the sets of inputs and labels in $d_t$ are denoted as $X_e^t$ and $Y_e^t$. 
However, vanilla LoRA tends to degrade the performance on previous edits due to catastrophic forgetting.
It's hence hard for the post-edit model to satisfy \textbf{Property 1$\sim$5}. In the following subsections, we present our MELO which overcomes this limitation with the cooperation of the vector database and dynamic LoRA modules.

\subsection{Sequential Editing with Dynamic LoRA}
Inspired by the prior work of DyLoRA \cite{dylora}, we explore to adapt dynamic LoRA to the sequential editing task, which can be well trained on partial ranks instead of the entire module. Unlike the original method that randomly select the range of LoRA ranks, we train non-overlapping LoRA blocks for different batches of edits to address the catastrophic forgetting problem.

As shown in Figure \ref{main}, we have low-rank matrices $B\in \mathds{R}^{m\times r}$ and $A\in \mathds{R}^{r\times d}$ for the LoRA module. Let's assume that we would like to train a part of weights in matrices $B$ and $A$ for each batch of edits, which can be termed as a trainable LoRA block. The range of a block is determined by the order number of a batch $t\in [1,n]$ and the predefined hyper-parameter partial rank $p$. In this way, the LoRA blocks for different batches of edits are non-overlapping:
\begin{equation}
\label{range}
    \begin{aligned}
        W_B^t &= B[\ :,\ (t-1)p:tp]\\
        W_A^t &= A[\ (t-1)p:tp,\ :\ ]
    \end{aligned}
\end{equation}
where $W_B^t$ and $W_A^t$ indicate the trainable block in the matrices $B$ and $A$ for the $t^{th}$ batch. The total rank equals to the number of needed LoRA blocks multiplied by the partial rank, thus MELO supports large editing batch size to keep less LoRA blocks. Table \ref{hyper} gives the default setting for MELO's training. 
With the learning rate $\eta$, a batch of edits $d_t$ can be quickly learned in a small LoRA block:
\begin{equation}
    \begin{split}
        W_B^{t} \leftarrow W_B^{t}-\eta \nabla_{W_B^t} \mathcal{L}[f(X_e^t;W_B^tW_A^t),Y_e^t]\\
        W_A^{t} \leftarrow W_A^{t}-\eta \nabla_{W_A^t} \mathcal{L}[f(X_e^t;W_B^tW_A^t),Y_e^t]
    \end{split}
\end{equation}
Since different batches of edits are trained with non-overlapping LoRA blocks, MELO could keep the information of previous edits without retraining. 

\subsection{Neuron Indexing with Vector Database}


In order to activate corresponding LoRA blocks for post-edit inputs during inference, we maintain an inner vector database (see Figure \ref{main}), which builds the neuron-index for editing samples as (key, value) pairs, where similar keys are clustered to represent the scope of associated editing samples, and values indicate the indexes of the LoRA blocks. For ease of understanding, we first introduce the components of our vector database. Then, we describe how to construct the cluster for the editing samples during training. After that, we explain how to locate the appropriate LoRA block by block searching in the inference stage.

\paragraph{Components:} 
During the training process, the vector database maintains the edit memories by building the neuron indexes, which contains following components:
\begin{itemize}
    \item \emph{Keys} ($K$): For each edit, the last hidden state $h^l$ obtained at layer $l$ is used as its key vector.  
    \item \emph{Values} ($V$): Each key maps to a value that represents the LoRA block index number.    
    \item \emph{Clusters} ($C$): Clusters contain the trained edits as key-value pairs. The keys in one cluster are close to each other by the Euclidean distance, and their average serves as the cluster center. 
    \item \emph{Radii} ($R$): Each cluster has a radius, which is changing during training to determine the editing scope.
\end{itemize}




\paragraph{Cluster Construction (Training Phase)}
For each edit, $(K,V)$ represents the key-value pair, $y_e$ is the target label and $C_{i^\star}$ indicates its nearest cluster with the radius $R_{i^\star}$. $R_{init}$ is a hyper-parameter for cluster initialization and update decision. $d(\cdot)$ measures the Euclidean distance of two input vectors. All situations during cluster construction are shown in Figure \ref{main}(c):

\begin{itemize}
    \item \textbf{Add:} If $d(K,C_{i^\star})\in(R_{i^\star} + R_{init},+\infty]$, a new cluster $\{C_e, [K:V], R_{init}, y_e\}$ can be initialized with the key itself as the center $C_e$.
    \item \textbf{Expand}: If $d(K,C_{i^\star})\in(R_{i^\star}, R_{i^\star}+ R_{init}]$ and the cluster label is same as the edit label, the cluster simply expands its radius to $d(K,C_{i^\star})$ to encompass this key, then add the $(K, V)$ pair into the cluster.
    \item \textbf{Conflict}: If $d(K,C_{i^\star})\in(R_{i^\star}, R_{i^\star}+ R_{init}]$ but the cluster label and the edit label are different, the radius of $C_{i^\star}$ will decrease and then a new cluster centered at $K$ with radius $d(K,C_{i^\star})/2$ will be added. Previous edits falling outside of $C_{i^\star}$ will be removed from the database.
\end{itemize}
Overall, the vector database maintains the clustered neuron indexes, where the keys can be efficiently searched during inference, and the corresponding values can be used to find certain LoRA blocks for editing. 


\paragraph{Block Searching (Inference Phase)} 
Given an input, we also use the last hidden state $h^l$ at layer $l$ as the query $K_q$. We first find the nearest cluster in the vector database, and then identify the closest key in this cluster.
\begin{equation}
\begin{split}
    i^{\star} &= \arg\min_{i} d(C_i, K_q), \forall C_i \in C\\
    j^{\star} &= \arg\min_{j} d(K_j, K_q), \forall K_j \in C_{i^{\star}}\\
\end{split}
\end{equation}
If $K_q$ falls in the radius of the nearest cluster $C_{i^\star}$, we map $i^\star$ and $j^\star$ to the LoRA block index with the value $V=C_{i^\star}[K_{j^\star}]$. After that, corresponding block matrices can be obtained based on Equation (\ref{range}) and the searched block index, which have been trained with similar editing samples and thus is appropriate for current editing.
If $K_q$ falls out of the radius of the nearest cluster, zero matrices are loaded as the LoRA block, thus the post-edit model uses original weights to infer the response (i.e., $\Delta{W}$ = 0 in Equation (\ref{lora_eq})). More concretely, the final block matrices used for editing can be formulated as:  
\begin{equation}
    W_BW_A = \{
    \begin{aligned}
        W_B^{V}W_A^{V}&,\ if\  d(C_{i^\star}, K_q) \leq R_{i^{\star}}\\
        \mathbf{0}\quad\   &,\ otherwise\\
    \end{aligned}
\end{equation}

\section{Experimental Setup}
\subsection{Datasets}


We perform extensive experiments on three well-known sequential editing tasks, including document classification, question answering and hallucination correction. The details about the datasets are described as follows:
\begin{itemize}[leftmargin=*, align=left]
\item \textbf{SCOTUS} is a subset of Fairlex \cite{fairlex}, which aims to categorize U.S.Supreme Court documents into 11 topics. Since the categorization rules change over time, the editor is required to correct realistic label shifts.
\item \textbf{zsRE} is a question answering (QA) dataset built upon zero-shot relation extraction \cite{zsre}. We split each QA pair and its rephrasings into two parts following previous studies \cite{serac,grace}, namely edits and holdouts. The holdouts are not directly edited during training, which are used to test the editing generality. A upstream dataset NQ \cite{NQ} is used to evaluate the locality. 
\item \textbf{Hallucination} is introduced by \cite{selfcheck} to correct the factual errors made by GPT models. 238 wikipedia-style biographies are generated by GPT-3, then 1392 sequential edits and 592 already-accurate outputs are created. The upstream dataset WebText \cite{webgpt} is used for testing the locality.
\end{itemize}

\subsection{Evaluation Metrics}
As described in prior studies \cite{mend,serac}, the most fundamental editing metrics are Edit Success (\textbf{ES}) and \textbf{Locality}, which are employed for all aforementioned datasets. In addition, we include two dataset-specific metrics. \textbf{Generality} \cite{rome,memit} is another essential property, and we quantify editors' generality on zsRE with the holdout dataset. For the Hallucination dataset, we additionally use the Accurate Attention Rate (\textbf{ARR}) for evaluating the performance on already-accurate outputs following previous studies \cite{grace}. We also report the editing speed and parameters for \textbf{Efficiency} study.

The evaluation functions vary for for different editing tasks. For document classification on SCOTUS, the average accuracy (\textbf{ACC}) is used \cite{fairlex}; Concerning question answering on the zsRE dataset, the mean F1 metric (\textbf{F1}) is applied \cite{grace}; Regarding to the hallucination correction task, we evaluate the performance of post-edit generative models through standard average perplexity (\textbf{PPL}) \cite{ppl}.
If $(x,y) \in D_{edit}$, the above measures stand for the \textbf{ES} metric. Similarly, they represent the \textbf{Locality} metric when $(x,y) \in D_{out}$. 





\subsection{Implementation Details}
The LLM backbones employed for editing vary on different datasets: BERT is used for the SCOTUS task; T5-Small and T5-Large are employed on the zsRE dataset; A pre-trained GPT2-XL is edited for the Hallucination correction.

Our proposed MELO is implemented based on the huggingface library PEFT\footnote{PEFT: https://github.com/huggingface/peft}, which can be easily integrated into multiple LLM backbones for model editing. 
Unless otherwise stated, the default hyper-parameter settings of MELO for different backbones are provided in Table \ref{hyper}. Detailed information about the location of layer for keys in vector database and the layer for integrating the dynamic LoRA modules are reported in the Appendix.

\begin{table}[thb]
\centering
    \resizebox{0.48\textwidth}{!}{
    \large
    \begin{tabular}{*{5}{l}}
    \toprule
      Hyper-param & BERT &  T5-Small & T5-Large& GPT2-XL  \\
        \midrule
        Partial Rank & 4 & 2 & 2 & 2 \\
        Initial Radius & 1.0 & 75.0 & 10.0 & 1.0  \\
        Batch Iteration & 40 & 30 & 40 & 50\\
        Learning Rate & $1e^{-3}$ & $1e^{-3}$ & $1e^{-3}$ &$1e^{-4}$ \\
        \bottomrule
    \end{tabular}
    }
    \caption{Default hyper-parameter settings of MELO.}
    \label{hyper}
\end{table}

\begin{table*}[t]
\renewcommand\arraystretch{1.1}
\normalsize
\centering
\begin{tabular}{lcccccccc}

    \toprule
    {} & \multicolumn{2}{c}{\textbf{SCOTUS (BERT; Acc $\uparrow$)}} & \multicolumn{3}{c}{\textbf{zsRE (T5-Small; F1 $\uparrow$)}} &\multicolumn{3}{c}{\textbf{Hal (GPT2-XL; PPL$\downarrow$)}} \\
    \cmidrule[0.7pt](rl){2-3}
    \cmidrule[0.7pt](rl){4-6}
    \cmidrule[0.7pt](rl){7-9}

        {Method} & {Locality} & {ES} & {Locality} & {ES} & {Generality} &  {Locality} & {ES} & {ARR}\\
    \midrule
    
    LoRA    &  0.21 & 0.16 & 0.33 & 0.26 & 0.15  & 2578.5 & 2187.6 & 1817.3 \\
    MEND    &  0.19 & 0.27 & 0.25 & 0.27 & 0.22 & 1369.8 & 1754.9 & 2902.5 \\
    SERAC   &  0.33 & 0.41 & 0.72 & 0.31 & 0.30 & 8183.7 & 133.3 & 10.04 \\ 
    CMR     &  0.52 & 0.52 & 0.56 & 0.82 & 0.74  & 1449.3 & 28.14 & 107.76 \\
    ROME    &  --- & --- & --- & --- & --- & 30.28 & 103.82 & 14.02 \\
    GRACE   &  0.81 & 0.82 & 0.69 & 0.96 & 0.94 & \textbf{15.84} & 7.14 & 10.00\\
    \midrule
    {MELO}   &  \textbf{0.96} & \textbf{0.92} & \textbf{0.72} & \textbf{0.98} & \textbf{0.97} & 17.45 & \textbf{1.04} & \textbf{2.66}\\
    \bottomrule
    \end{tabular}
    \caption{Comparison results of MELO and the recent model editing methods on various sequential editing tasks. } 
\label{results}
\end{table*}

\subsection{Baselines}
We compare our proposed MELO with recent advanced baselines: 1) Vanilla \textbf{LoRA} \cite{lora} is a typical parameter-efficient tuning method, which integrates low-rank adapters to target modules and only updates these adapters during sequential editing; 2) \textbf{MEND} \cite{mend} learns a hyper-network with additional training data to transform the gradient obtained by standard fine-tuning; 3) \textbf{SERAC} \cite{serac} decomposes editing into three sub-models and  additionally trains the scope classifier and counterfactual model, which routes the inputs to alter the model's behavior; 4) \textbf{ROME} \cite{rome} locates knowledge in specific layers of GPT and directly modify these layers for extensive edits. Since ROME is especially designed for GPT models, it is only involved in the Hallucination task; 5) \textbf{CMR} \cite{cmr} is a method based on continually learning, which fine-tunes the input model sequentially to output a refined model for processing future examples; 6) \textbf{GRACE} \cite{grace} replaces the hidden states of in-scope inputs with pre-trained parameters according to an edit codebook.

\section{Results and Analyses}

\subsection{Main Results}
Table \ref{results} shows the results of the recent baselines and our proposed method. We observe that our MELO is significantly superior to the exiting editing methods without using any additional training data.  
Specifically, we outperform the recent advanced baseline GRACE by up to $15\%$ improvements regarding to the Local and ES metrics in most cases, indicating the effectiveness of our method in accurately altering models’ behavior for the editing samples without interference on others.  
In addition, we also achieve significant improvements in terms of Generality on zsRE, which demonstrates that our method is effective in editing for more associated samples that are similar to the training stage. 
For the Hallucination task with the 1.5B GPT2-XL backbone, our MELO achieves the overwhelming advantages on ES and ARR, and performs slightly worse for the Local metric compared with Grace, which further certifies that our method could efficiently edit the large-scale model and well retains the performance on the originally accurate inputs. 


\subsection{Efficiency of Editing}

We compare the efficiency of editing with the recent advanced baselines including SERAC and GRACE. The former is a representative memory-based editor, while the latter is the existing best editing method on sequential editing tasks. With a single Nvidia RTX 3090 GPU, we investigate the editing speed and the amount of extra parameters used on zsRE dataset.

\begin{table}[htb]
\renewcommand\arraystretch{1.3}
\normalsize
\centering

    \resizebox{0.48\textwidth}{!}{
    \begin{tabular}{llllll}

    \toprule
    {} & \multicolumn{2}{c}{T5-Small (60M)} & \multicolumn{2}{c}{T5-Large (770M)} \\
    \cmidrule[0.7pt](rl){2-3}
    \cmidrule[0.7pt](rl){4-5}

    {Method} & {Speed } & {Param}  & {Speed} & {Param} & {Num} \\
    \midrule
    SERAC   & --- & 126M   & --- & 126M & 1k \\
    GRACE   & 47.55 edits/min & 0.51M   &7.422 edits/min & 1.02M & 1k \\
    MELO    & 2464 edits/min  & 0.12M  & 401.6 edits/min   & 0.41M & 1k\\

    \bottomrule
    \end{tabular}
    }
    \caption{Efficiency of editing on zsRE.}
\label{resource}
\end{table}

As shown in Table \ref{resource}, we observe that MELO needs the least extra parameters to perform model editing, since a batch of edits only requires $1$ block of dynamic LoRA with low partial rank. For example, if editing $1k$ inputs for the T5-Small model, using the batch size of $100$ and partial rank of $2$, with $4$ linear layers incorporated with dynamic LoRA, the total extra parameters would then be:
\begin{equation}
    0.12M \approx 4 * (1k/100) * [(1024 * 2 + 2 * 512)]
\end{equation}
where $1024$ and $512$ are the input and output dimension in the linear layer. While GRACE needs to train a $512$-dimension vector for each edit, and SERAC routes edits among three sub-models, which results in large amount of extra parameters. In addition, GRACE edits model in a sequential manner with the batch size of $1$, which requires much more editing time. In particular, our editing speed is more than 50 times of GRACE. The editing speed of SERAC is not presented, since it needs additional training on two extra models (scope classifier and counterfactual model).







\begin{figure}[htb]
\centering

\subfigure[Cluster Number]{
\label{cluster}
\includegraphics[width=0.43\columnwidth]{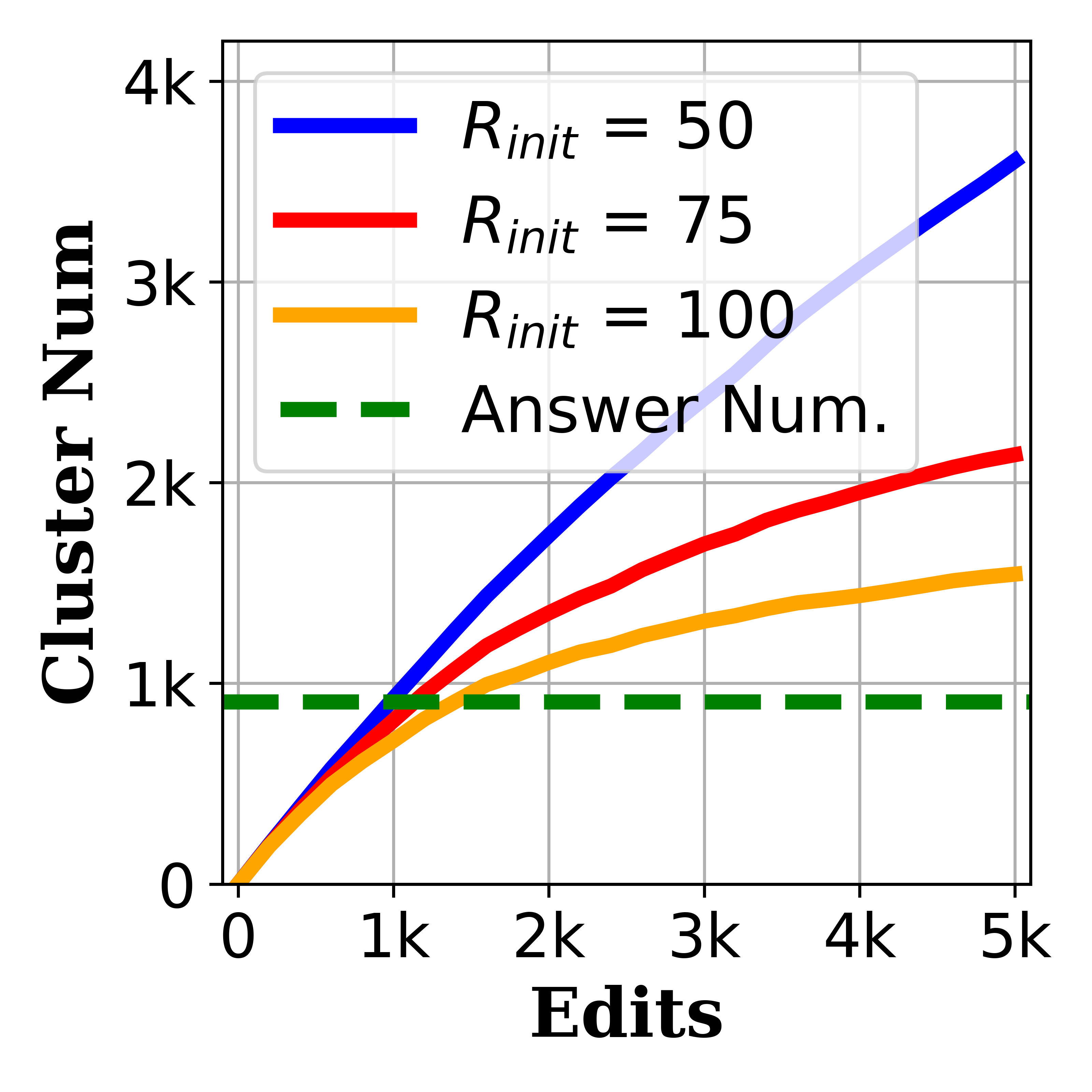}
}
\subfigure[Conflict Number]{
\label{conflict}
\includegraphics[width=0.43\columnwidth]{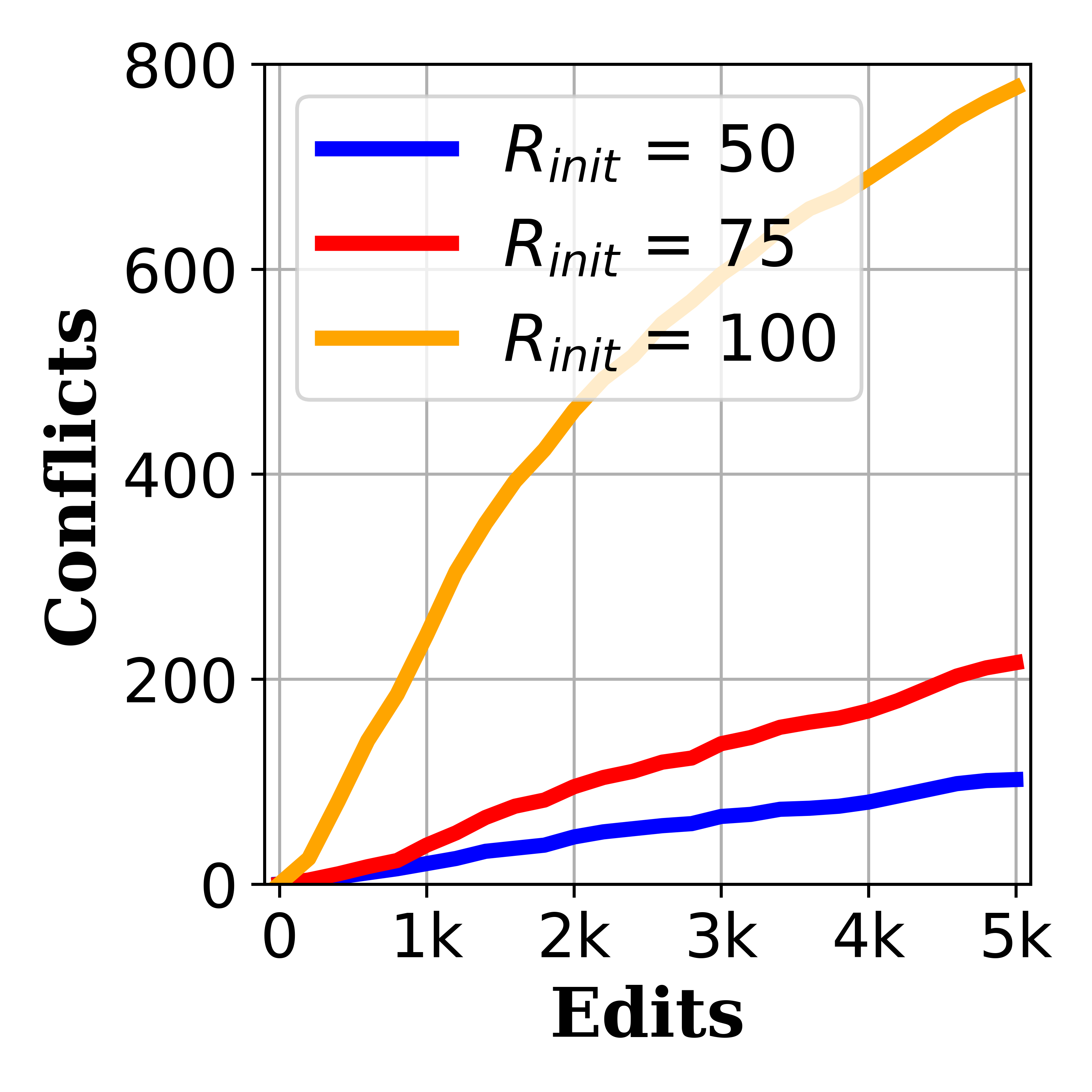}
}
\subfigure[Key Visualization]{
\label{PCA}
\includegraphics[width=0.43\columnwidth]{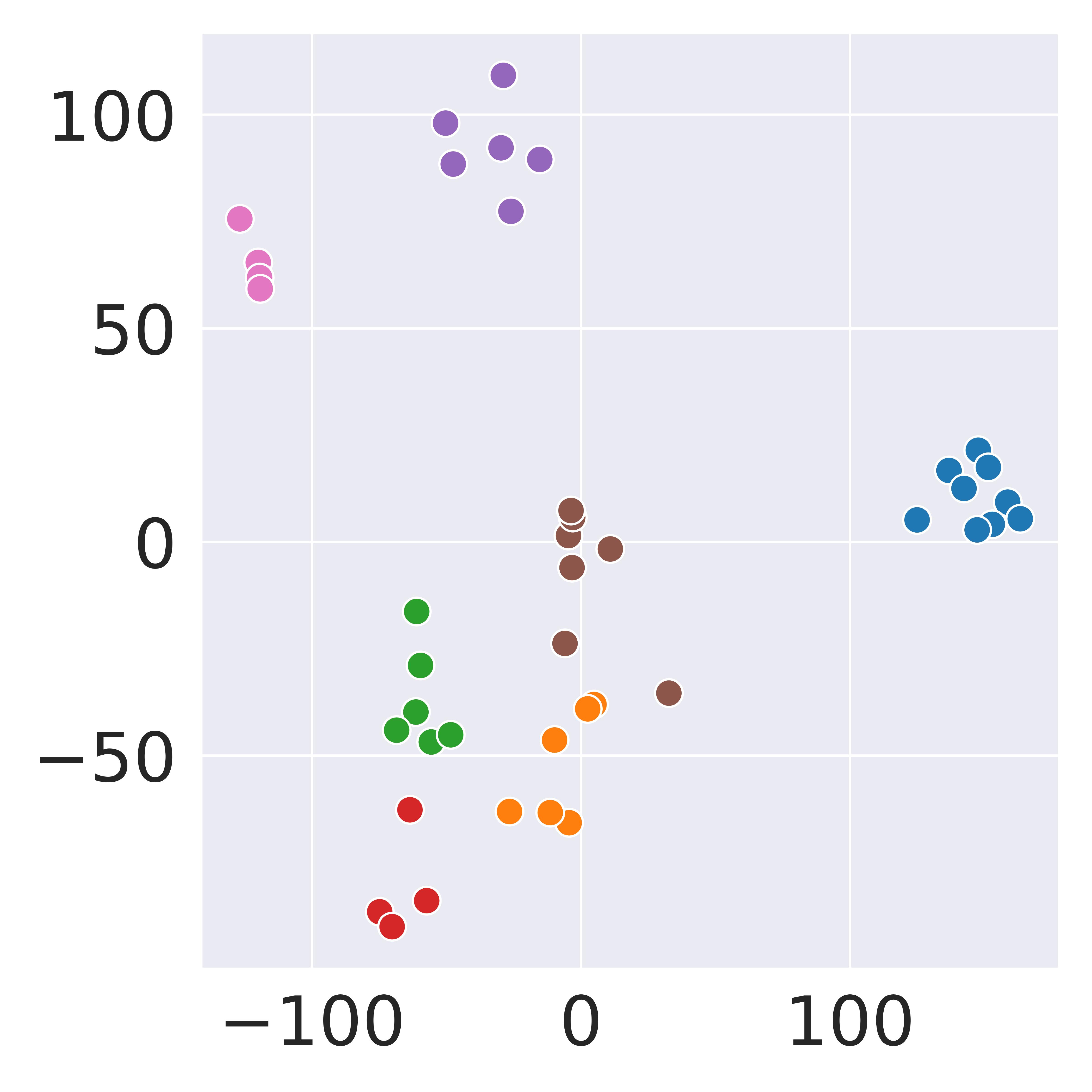} 
}
\subfigure[Forgotten Number]{
\label{forget}
\includegraphics[width=0.43\columnwidth]{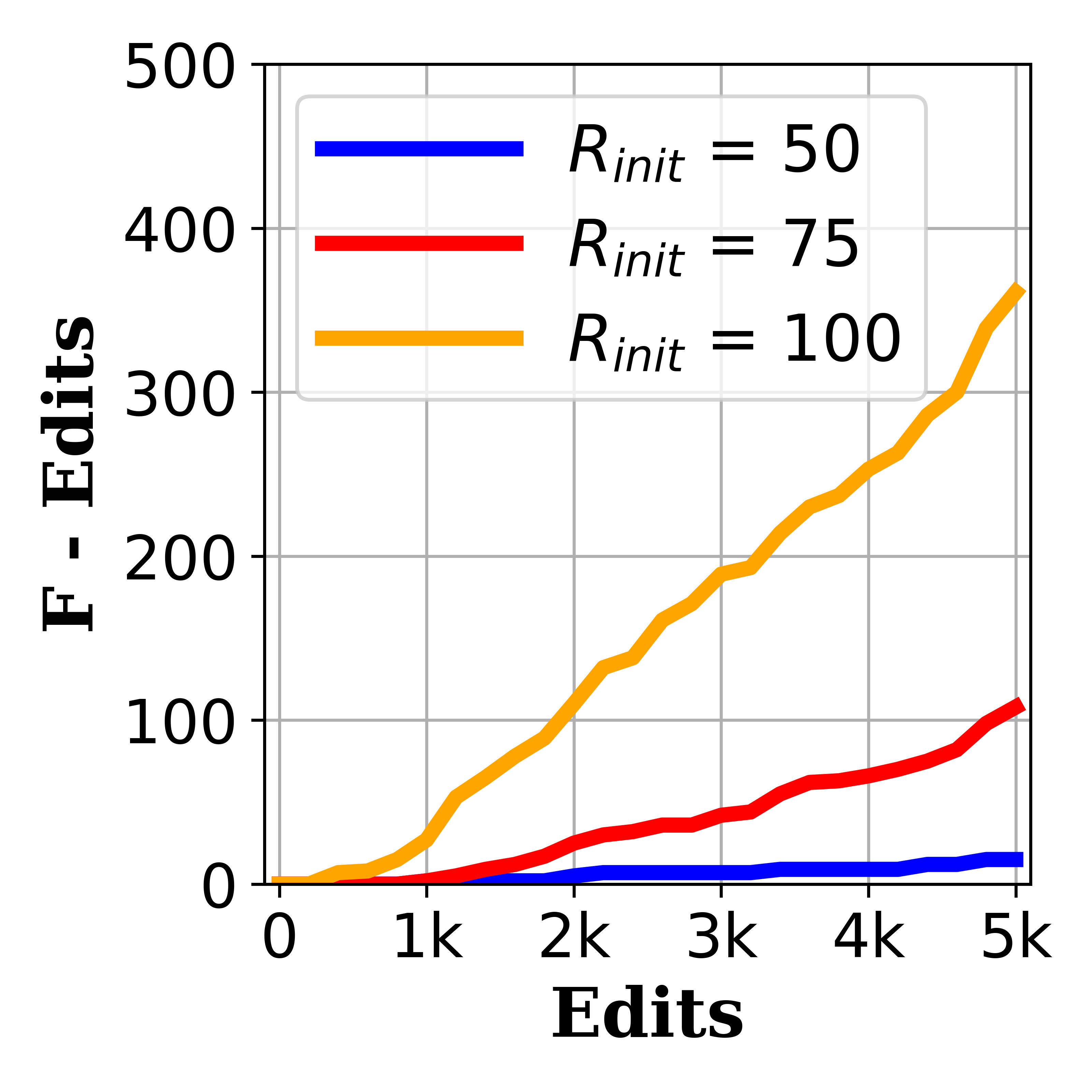}
}
\caption{Effect of initial cluster radius $R_{init}$ on zsRE.}
\label{radius}
\end{figure}

\begin{figure*}[h]
\centering
\subfigure[Backbone: \textbf{T5-Large} ]{
    \centering
    \includegraphics[width=0.90\textwidth]{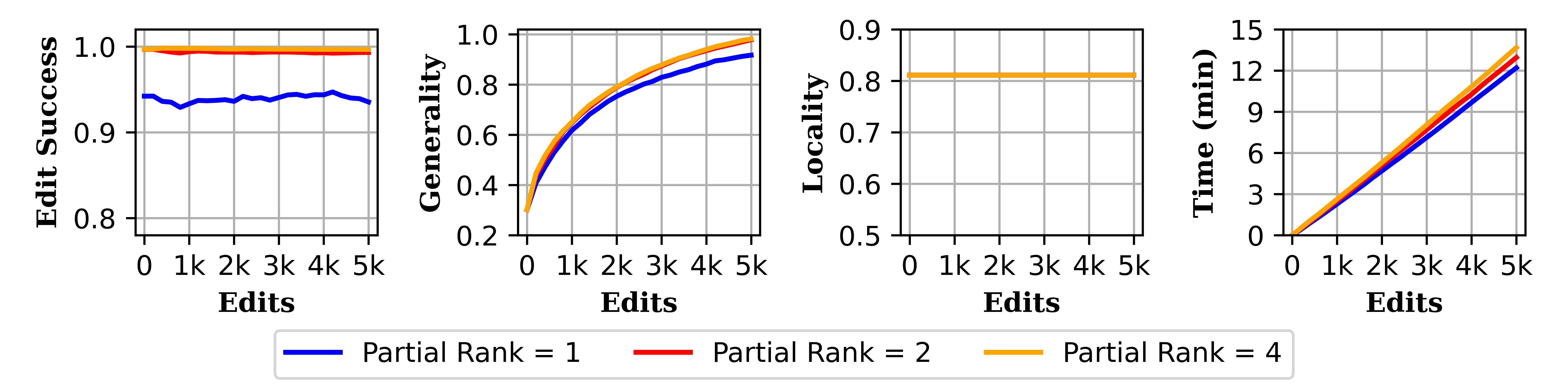}
}
\subfigure[Backbone: \textbf{T5-Small} ]{
    \centering
    \includegraphics[width=0.90\textwidth]{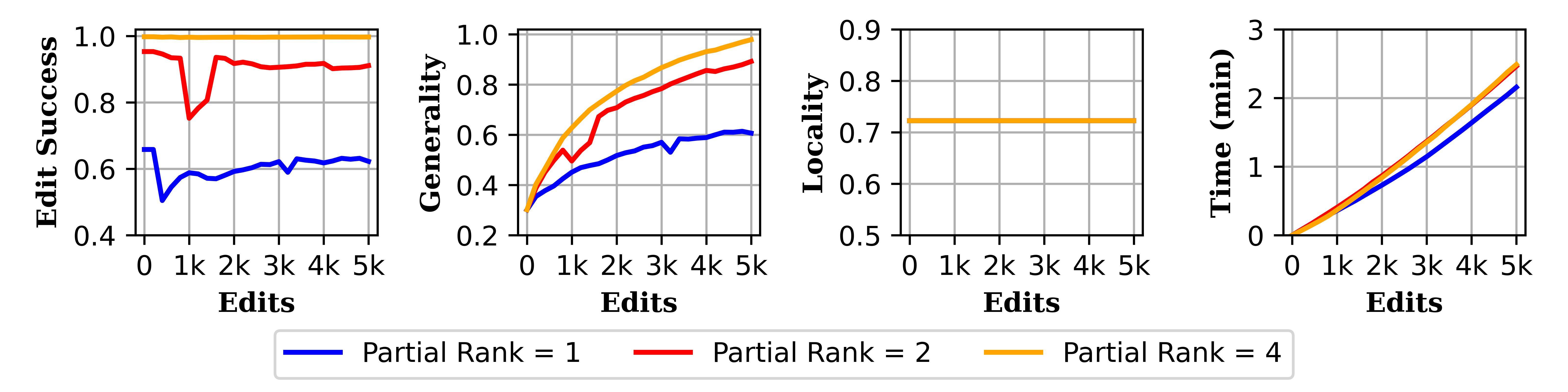}
}

\caption{Effect of the partial rank of dynamic LoRA on zsRE. }
\label{rank}
\end{figure*}

\subsection{Further Analyses of MELO}

\paragraph{Effect of Cluster Radius.} 
We perform a set of experiments to study how the initial cluster radius affects the neuron-index construction during editing. For limited space, we only present the results on zsRE dataset in Figure \ref{radius}, where $R_{init}$ varies in $\{50, 75, 100\}$. Similar results can be observed on other datasets.
As PCA shown in Figure \ref{PCA}, the keys of rephrasings belonging to the same question are close to each other in the semantic space, which serves as a warranty to accurately identify the editing scope in the inference stage. The influence of different cluster radii are shown in Figure \ref{cluster}, \ref{conflict} and \ref{forget}. 
Ideally, the cluster number should be equivalent to the number of answers with multiple question rephrasings.
We observe that using larger cluster radius is helpful to decrease the total number of clusters, and therefore alleviate the computation cost of LoRA block indexing in the vector database. Whereas, increasing the radius will also provoke more cluster conflicts, which consequently lead to more forgotten edits. In our experiments, we recommend a reliable setting as described in Table \ref{hyper} for $R_{init}$.

\paragraph{Effect of Partial Rank of Dynamic LoRA.}
The partial rank of a LoRA block determines how many neurons are used to learn a batch of edits. To investigate its effect on the editing performance, we evaluate MELO with different partial ranks on zsRE based on two language models (T5-Small and T5-Large), with each block trained on $100$ edits. 
The results are shown in Figure \ref{rank}. We observe that larger partial ranks usually result in better performance in edit success and generality, which is more evident with the smaller language model T5-Small. This corresponds to our intuition that when using larger partial ranks, more neurons are incorporated to learn and store the editing knowledge, which consequently improves the editing performance.
It is also interesting to find that the performance on locality remains the same with various partial ranks, since our vector database is effective to identify the editing scope, and no LoRA blocks are invoked for the out-of-scope samples.
For the time cost, there are no significant differences with various partial ranks, since only a few neurons are used for learning, which is highly efficient. 

\begin{figure}[htb]
\centering
\includegraphics[width=0.90\columnwidth]{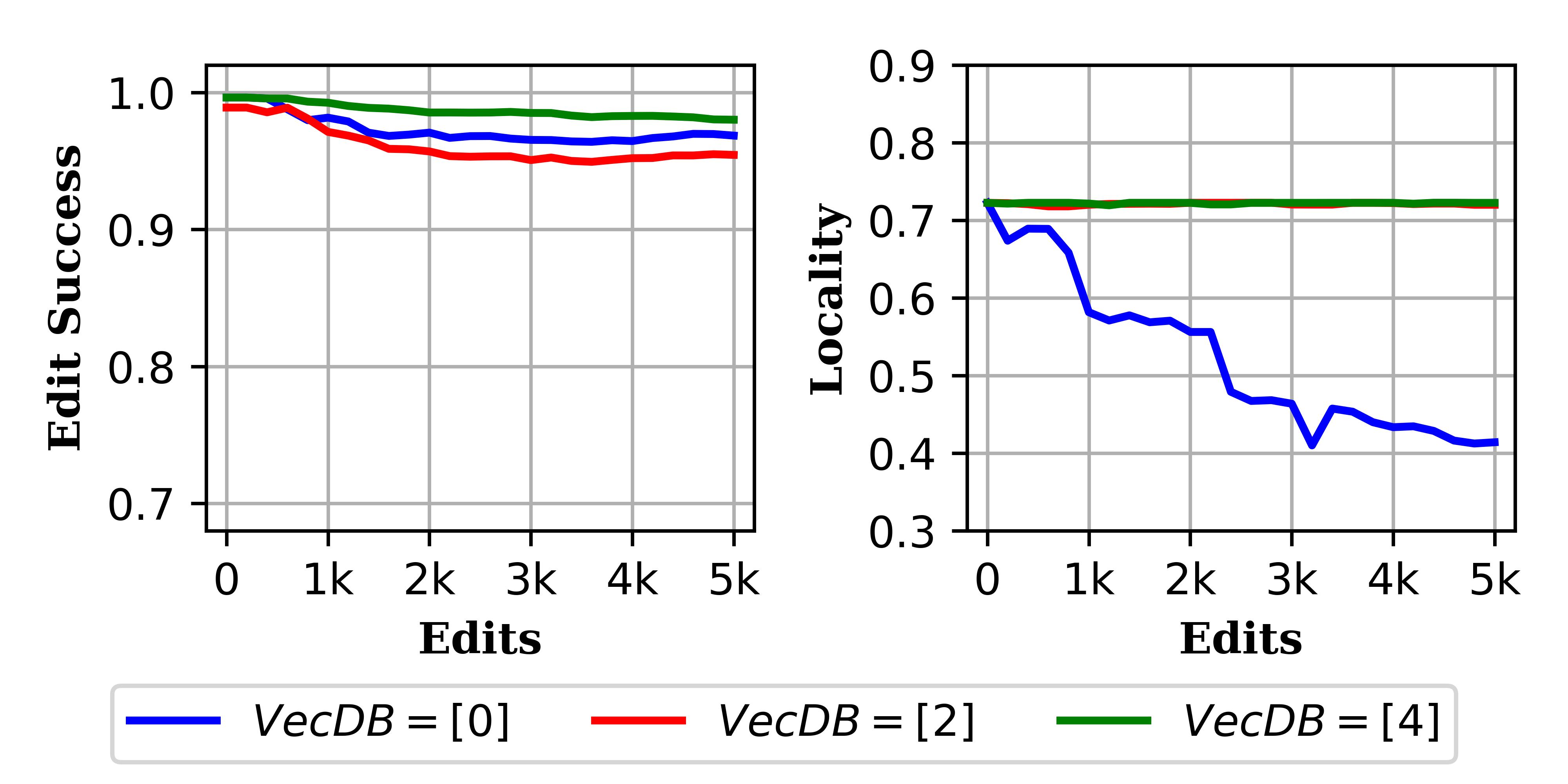}
\caption{Effect of using different layers for key representation in the vector database.}
\label{block}
\end{figure}

\paragraph{Effect of Key Layer for Vector Database.}
To study the impact of using the hidden state in different neural layers as keys for the vector database, we experiment with T5-Small on zsRE varying the layers in $\{0, 2, 4\}$. As illustrated in Figure \ref{block}, keys based on the fourth layer achieve the best performance in terms of edit success and locality. 
In addition, there are slight differences in edit success when using different layers as keys. While regarding to the locality, the performance decreases dramatically when using the first layer for keys, indicating the poor ability in editing scope identification and thus intervenes the out-of-scope samples during editing. 
This observation is in line with the findings in prior work \cite{kv_memory} that shallow layers can only detect the shallow sentence patterns, while the upper layers encode more semantic features. Therefore, except the first layer, any upper layer (prior to LoRA modules) can be used for keys, which yields better editing performance in our experiments.

\section{Conclusions}
In this paper, we propose a novel method for sequential model editing, which dynamically activates the corresponding LoRA blocks indexed in an inner vector database to alter the behaviour of models. Extensive experiments on three editing tasks confirm that our method outperforms the recent advanced baselines on various editing metrics. It is also notable that our method shows great advantages in editing efficiency, with 50 times faster editing speed of the best baseline.
In the future, we will explore more effective neuron-indexed vector database, and extend MELO to more scenarios such as multi-modal model editing.

\newpage
\section{Acknowledgements}
This research is funded by the National Key Research and Development Program of China (No. 2021ZD0114002), the National Natural Science Foundation of China (No. 62307028), and the Science and Technology Commission of Shanghai Municipality Grant (No. 22511105901, No. 21511100402, No.23ZR1441800).

\bibliography{aaai24}

\appendix
\clearpage

\setcounter{figure}{0}
\setcounter{section}{0}
\renewcommand{\thefigure}{A\arabic{figure}}
\setcounter{table}{0}
\renewcommand{\thetable}{A\arabic{table}}

\section{Appendix}

\subsection{A1. Default Location Settings}
With editing tasks based on different LLM backbones, the location of layer for keys in vector database and the layer for integrating the Dynamic LoRA modules can be treated as hyper-parameters. Following tables demonstrate the default location settings in our experiments (Table \ref{A1} for \textbf{BERT}, Table \ref{A2} for \textbf{T5-Small}, Table \ref{A3} for \textbf{T5-Large} and Table \ref{A4} for \textbf{GPT2-XL}). 

\begin{table}[H]
\centering
\begin{tabular}{l} 
\toprule
Layer with Vector Database\\
\midrule
\small{\texttt{bert.encoder.layer[4].output.dense}}\\
\midrule
Dynamic LoRA Target Modules\\
\midrule
\small{\texttt{bert.encoder.layer[9].output.dense}}\\
\small{\texttt{bert.encoder.layer.[10].output.dense}}\\
\small{\texttt{bert.encoder.layer.[11].output.dense}}\\
\bottomrule 
\end{tabular}
\caption{Default MELO Target Modules for \textbf{BERT}} 
\label{A1}
\end{table}

\begin{table}[H]
\centering
\begin{tabular}{l} 
\toprule
Layer with Vector Database\\
\midrule
\small{\texttt{Encoder.Block[4].Layer[1].DenseReluDense.wo}}\\
\midrule
Dynamic LoRA Target Modules\\
\midrule
\small{\texttt{Encoder.Block[5].Layer[1].DenseReluDense.wo}}\\
\small{\texttt{Decoder.Block[5].Layer[2].DenseReluDense.wo}}\\
\small{\texttt{Encoder.Block[6].Layer[1].DenseReluDense.wo}}\\
\small{\texttt{Decoder.Block[6].Layer[2].DenseReluDense.wo}}\\
\bottomrule 
\end{tabular}
\caption{Default MELO target modules for \textbf{T5-Small}}  
\label{A2}
\end{table}

\begin{table}[H]
\centering
\begin{tabular}{l} 
\toprule
Layer with Vector Database\\
\midrule
\small{\texttt{Encoder.Block[4].Layer[1].DenseReluDense.wo}}\\
\midrule
Dynamic LoRA Target Modules\\
\midrule

\small{\texttt{Encoder.Block[22].Layer[1].DenseReluDense.wo}}\\
\small{\texttt{Decoder.Block[22].Layer[2].DenseReluDense.wo}}\\
\small{\texttt{Encoder.Block[23].Layer[1].DenseReluDense.wo}}\\
\small{\texttt{Decoder.Block[23].Layer[2].DenseReluDense.wo}}\\
\bottomrule 
\end{tabular}
\caption{Default MELO Target Modules for \textbf{T5-Large}} 
\label{A3}
\end{table}

\begin{table}[H]
\centering
\begin{tabular}{l} 
\toprule
Layer with Vector Database\\
\midrule
\small{\texttt{transformer.h[35].mlp.c\_fc}}\\
\midrule
Dynamic LoRA Target Modules\\
\midrule
\small{\texttt{transformer.h[36].mlp.c\_fc}}\\
\small{\texttt{transformer.h[37].mlp.c\_fc}}\\
\bottomrule 
\end{tabular}
\caption{Default MELO Target Modules for \textbf{GPT2-XL}}  
\label{A4}
\end{table}
Note that the vector database is integrated in the layer prior to all dynamic LoRA modules in
different LLMs, this is because the indexing for LoRA blocks should be done before the forward pass of LoRA modules.
\subsection{A2. Explanation for the PCA Analysis}
Figure \ref{PCA} visualizes the result of principal component analysis (PCA) on edit keys, which are close to each other if the edits have similar semantics. Each cluster contains sentences with the same target label. Examples are shown in Table \ref{pca_table}:
\begin{table}[h]
\centering
\begin{tabular}{l} 
\midrule
\tikz\draw[orange,fill=orange] (0,0) circle (.4ex); Answer: Crater\\
\midrule
What is the constellation that HD 98800 is a part of ?\\
To which constellation does HD 98800 belong ?\\
\midrule
\tikz\draw[purple,fill=purple] (0,0) circle (.4ex); Answer: National Geographic Channel\\
\midrule
In what network is Beast Hunter ?\\
Which broadcasting company broadcast Beast Hunter ?\\
\midrule 
\tikz\draw[blue,fill=blue] (0,0) circle (.4ex); Answer: Draco\\
\midrule
Which was the constellation for HD 151613?\\
Which is the constellation of HD 151613?\\
\midrule
\tikz\draw[brown,fill=brown] (0,0) circle (.4ex); Answer: 4 April 1960 \\
\midrule
The point in time of 32nd Academy Awards was what ?\\
What was the date of the 32nd Academy Awards ?\\
\midrule

\end{tabular}
\caption{Example Sentences of each Cluster in Figure \ref{PCA}}  
\label{pca_table}
\end{table}

\subsection{A3. Dataset for Generality Evaluation}
As described in the \textcolor{blue}{Introduction} and \textcolor{blue}{Problem Formulation}, the Property 3 \emph{Generality} is used to measure whether the post-edit model can make correct predictions for the paraphrased inputs, which are unseen during editing but closely associated with the intended edits. Here is an example:

\begin{table}[h]

\normalsize
\centering

\begin{tabular}{l} 
\midrule
\tikz\draw[pink,fill=pink] (0,0) circle (.4ex); Answer: Purple Mountain Observatory\\
\midrule
\rowcolor{table_yellow} \texttt{Intended Edit}\\
Who discovered the 2752 Wu Chien hit ?\\
\rowcolor{table_blue} \texttt{Holdouts}\\
By whom were 2752 Wu Chien-shiung discovered ?\\
Who was the explorer of 2752 Wu Chien-Shiung ?\\
\midrule
\end{tabular}
\caption{Intended edits and rephrased holdouts.}  
\label{example}
\end{table}
To evaluate the generality on zsRE described in \textcolor{blue}{Datasets}, we split each QA pair and its rephrasings into two groups with equal amount, namely intended edits and holdouts. In \textcolor{blue}{Figure 4}, the F1-Score curves demonstrate well generality of our method on the entire holdout set with increasing number of completed edits. \textcolor{blue}{Table 2} shows better generality of our MELO compared to other editors, which benefits from the accurate editing scope built in the vector database.
\end{document}